\documentclass{INTERSPEECH2023}
\usepackage{cite}
\usepackage{amsmath}
\usepackage{amssymb}
\interspeechcameraready

\title{Leveraging Label Information for Multimodal Emotion Recognition}
\name{Peiying Wang, Sunlu Zeng, Junqing Chen, Lu Fan, Meng Chen\sthanks{Corresponding author.}, Youzheng Wu, Xiaodong He}
\address{ JD AI Research, Beijing, China}
\email{\{wangpeiying3,zengsunlu1,chenjunqing3,fanlu,chenmeng20,wuyouzheng1,xiaodong.he\}@jd.com}
\usepackage{threeparttable}
\begin{document}

\maketitle

\begin{abstract}
Multimodal emotion recognition (MER) aims to detect the emotional status of a given expression by combining the speech and text information. Intuitively, label information should be capable of helping the model locate the salient tokens/frames relevant to the specific emotion, which finally facilitates the MER task. Inspired by this, we propose a novel approach for MER by leveraging label information. Specifically, we first obtain the representative label embeddings for both text and speech modalities, then learn the label-enhanced text/speech representations for each utterance via \textit{label-token} and \textit{label-frame} interactions. Finally, we devise a novel label-guided attentive fusion module to fuse the label-aware text and speech representations for emotion classification. Extensive experiments were conducted on the public IEMOCAP dataset, and experimental results demonstrate that our proposed approach outperforms existing baselines and achieves new state-of-the-art performance.

%Multimodal emotion recognition(MER) aims to extract emotion related information from multiple sources and integrate different modal representations for sentiment analysis. 
%Speech emotion recognition (SER) is a challenging problem since human convey emotions in subtle and complex ways, and extracting the emotion related information from multiple sources is the real crux of the matter. 
%Recent works have witnessed impressive improvements due to the benefits of pre-trained models. However, the existing approaches treat labels as independent and meaningless one-hot vectors, which cause a loss of potential label information for extracting emotional information.In this work,  we propose a novel label embedding enhanced attentive framework for MER. Specifically, the model makes full use of both textual label and speech/tonal label information, where the \textit{label-word} and \textit{label-frame} attention based target will be introduced to locate the emotional-relevant words and frames, respectively. In addition, we consider that the shared label space can serve as a bridge to efficiently integrate the information of two modalities, thus we devise a label-centered cross attention module to align the text words with the speech frames from the label correlation perspective, %aiming to produce more accurate multimodal feature representations. which is then used to fuse emotional features from two modalities. We evaluate the approach on the IEMOCAP dataset and the experimental results show the proposed approach achieves the state-of-the-art performance on the dataset.

\end{abstract}

\section{Introduction}

% Human express emotion through various modalities such as voice, facial expression, body posture, therefore utilizing multiple modalities may accurately capture expressed emotion and lead to better recognition results than unimodal approaches. Latest researches \cite{chen2020multi,morais2022speech,priyasad2020attention,xi2022frontend,yoon2018multimodal,krishna2020multimodal,wu2021emotion} have also proved that multimodal methods outperform the unimodal methods. Consequently, multimodal emotion recognition(MER) has been a hot research topic in recent years. 
% Generally, the emotion of spoken language is beyond the content of the utterance itself and is also related to the speaker's tone.
% Therefore, to completely understand the emotion of the speaker, it is not enough to solely rely on what he said, but also needs to consider the tone he expressed with.
% To simulate this phenomenon, the text- and speech-based multimodal emotion recognition (MER) task was proposed~\cite{busso2008iemocap}.
Generally, the emotion of spoken language is beyond the linguistic content of the utterance itself, and it is also related to the speaker's voice characteristics.
To completely understand the emotion of the speaker, the text- and speech-based multimodal emotion recognition (MER) task was proposed to identify the emotion within an utterance~\cite{busso2008iemocap}.

\begin{figure}[!tb]
  \centering
  \includegraphics[width=0.8\linewidth]{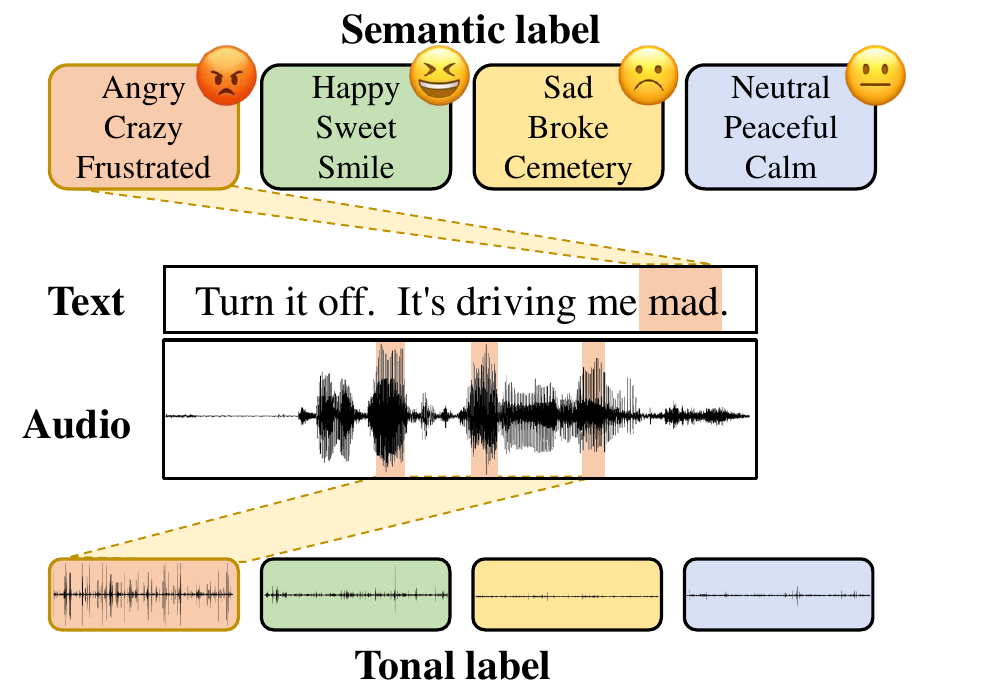}
  \caption{Visualization of labels. %The \textbf{semantic label} represents the set of the most emotion relevant words for each class, and \textbf{tonal label} is produced by concatenating the waveforms of all the key-frames for each class.
  The \textbf{semantic label} presents the emotion relevant words for each class, and the \textbf{tonal label} displays the waveforms generated by concatenating the key-frames under each class.
  }
  \label{fig:label_visualization}
  \vspace{-2.0em}
  \setlength{\abovecaptionskip}{-1.em}
\end{figure}

Recently, MER has attracted more and more attention.
% % Benefiting from the rapid development of deep learning, existing work on this task has evolved from rule-based to neural network-based methods.
% % For example, Yoon et al.~\cite{DBLP:conf/slt/0002BJ18} proposes to adopt dual recurrent neural networks to learn the information from audio and text sequences and combine them to predict the emotion.
% % Peri et al.~\cite{DBLP:conf/icassp/PeriPBS21} combine audio and video information and utilize multitask setting for emotion recognition.
% Benefiting from the development of deep learning, existing work mainly concentrated on neural network-based methods.
% For example Yoon et al.~\cite{DBLP:conf/slt/0002BJ18} and Peri et al.~\cite{DBLP:conf/icassp/PeriPBS21} attempt to apply recurrent or convolution neural networks to learn the semantic representation for the text and speech sequences.
% Recently, with the advancement of the Pre-training language models (PLM), e.g. BERT  \cite{devlin2018bert} and wav2vec \cite{baevski2020wav2vec}, some work attempt to apply them on this task.
% For instance, Li et al.\cite{li2022context} proposed a context-aware multimodal fusion framework for the MER task, which apply BERT and WarLM as encoders.
% Chen et al.\cite{chen2022key} proposed a key-sparse Transformer based on the RoBERTa and Wav2vec, and it focuses more on emotional-related information.
In the early phase, most works explored rule-based and neural network-based methods~\cite{DBLP:conf/slt/0002BJ18,DBLP:conf/icassp/PeriPBS21}.
With the rapid development of self-supervised learning and pre-training, researchers attempt to tackle this task based on pre-trained models, e.g. BERT  \cite{devlin2018bert} and wav2vec2.0 \cite{baevski2020wav2vec}.
%advancement of the Pre-training language models (PLM), e.g. BERT  \cite{devlin2018bert} and Wav2vec \cite{baevski2020wav2vec}, some work attempts to apply them to this task.
For instance, Li et al.\cite{li2022context} proposed a context-aware multimodal fusion framework for the MER task, which applied BERT and WavLM as encoders.
Chen et al.\cite{chen2022key} proposed a key-sparse Transformer based on the RoBERTa and Wav2vec, which focuses more on emotion-related information.
Despite their success, most of them only take labels as supervised signals while neglecting their inherent semantic information.
% Intuitively, the label information is capable of helping the model to better understand the utterance and then improving the performance of the model.
Intuitively, the label information should be capable of helping the model to better understand the utterance.
% As shown in Figure~\ref{fig:label_visualization}, different emotion corresponds to different set of salient tokens and frames.
% Based on this character, the model is able to locate the task-oriented salient tokens/frames accurately under the guidance of the label information.
As shown in Figure~\ref{fig:label_visualization}, for the text input, the token ``mad'' is similar to the label \textit{angry} in semantics.
As to the speech input, some frame segments also have in common with the tonal label.
Based on above observation, we argue that the model may be able to locate the task-oriented salient tokens/frames accurately under the guidance of the label information.
Then, the model can pay more attention to the key information and effectively ignore the interference of redundant information.
Therefore, as a kind of prior knowledge, leveraging label information is essential for the MER task.

Label embedding is to learn the embeddings of the labels in classification tasks and has been proven to be effective in computer vision and natural language processing ~\cite{wang2018joint,xiong2021fusing,zhang2022label,le2022legal}, which enjoys a built-in ability to leverage alternative sources of information related to labels, such as class hierarchies or textual descriptions.
%The application of label information has been explored by previous works in the fields of natural language processing and computer vision~\cite{wang2018joint,xiong2021fusing,zhang2022label,le2022legal}.
However, there are rare speech-related work devoted to this technology. Take MER for example, there exist at least two obstacles that need to resolve.
%We take the MER as an example to analyze the reason behind that, and come out with two points.
Firstly, labels are usually in the form of text.
Due to the inherent disparities between the speech and the text, they cannot be directly exploited in speech-related tasks. How to obtain representative label embeddings for the speech modality becomes a big challenge. 
Secondly, when introducing the label information, it will increase the difficulty of multimodal fusion. How to project the text/speech representations into the same label embedding space and fuse the multimodal features seamlessly is also a critical issue.
%Thereby, to introduce the label information into speech-related work, we face two challenges: 1) how to learn the label representation in the speech, 2) how to fuse the label-aware multimodal representation.

%In this work, we propose a novel label enhanced model for multimodal emtion recognition (LE-MER). Specifically, for the text modality, we leverage the textual/semantic label information to locate the relevant emotional tokens. To this end,  a \textit{label-word} attention based objective function will be introduced to encourage the emotional-relevant words are weighted higher than the irrelevant ones. As for the speech modality, the corresponding textual label cannot be directly used/exploited due to the inherent disparities between the two modalities. Thus, we select some representative audio frames as the tonal-label for each emotion category, then a similar objective will be introduced based on the \textit{label-frame} attention matrix to help the model pay attention to the emotion related frames. In addition, we further design a label-centered cross-attention mechanism to align the speech frames and text words by the emotion-related information, where the shared label space will serve as the bridge to integrate the emotional information from multiple cues more effectively.

 \begin{figure*}[!htb]
 \vspace{-1.0em}
  \centering
  \includegraphics[width=\linewidth]{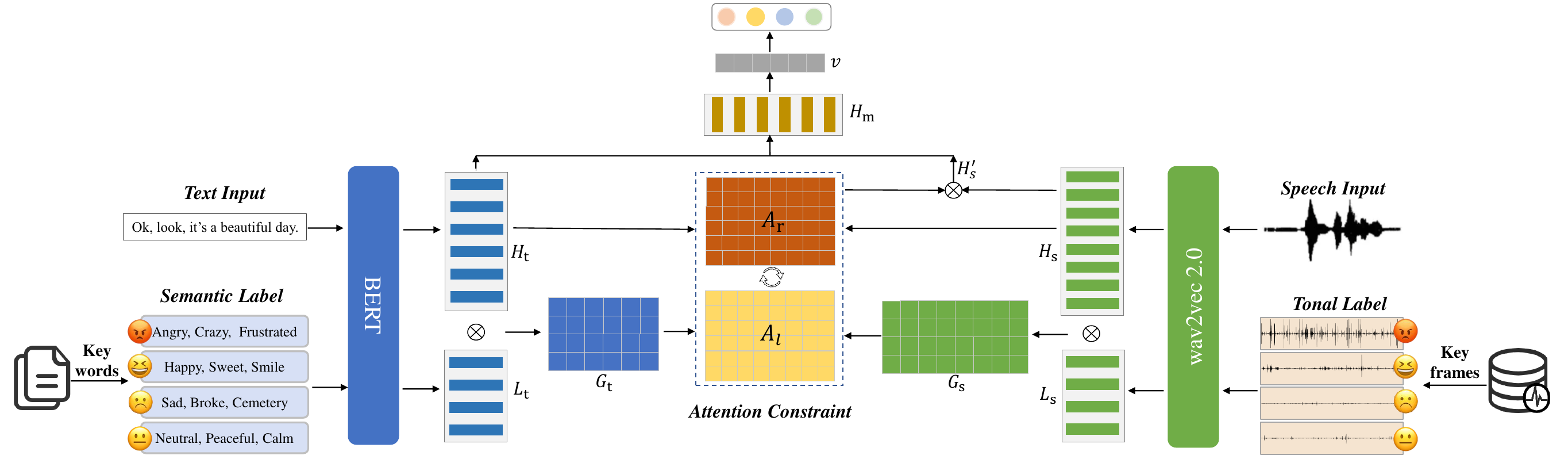}
  \caption{The architecture of our proposed model LE-MER.}
  \label{fig:speech_production}
  \vspace{-2.0em}
\end{figure*}

% In this work, we propose a novel label-enhanced model for multimodal emotion recognition (LE-MER) to tackle the above problems.
% Formally, we first extract salient tokens/frames from the whole dataset for each class as their extensions.
% Namely, each kind of label information consists a set of tokens and frames.
% After that, we integrate them into the model to assist the prediction of each utterance.
% Specifically, we adopt BERT and wav2vec to encoder the text and speech in the given utterance and the label information respectively.
% Then, we leverage the label information to locate the salient tokens/frames in the utterance via the cross-attention mechanism.
In this work, we propose a novel framework for MER to tackle the above challenges.
We first summarize representative tokens/frames from the training set for each class as their descriptions.
% Namely, each kind of label information consists a set of tokens and frames.
For text, we extract salient words for each label based on the frequency of tokens.
As for speech, we utilize wav2vec2.0 to discretize the whole dataset, and then extract salient frames from them.
After obtaining that, we further devise a novel label-enhanced multimodal emotion recognition model (LE-MER).
% to leverage the label information to assist the prediction of each utterance.
Formally, given an utterance and the extracted label information, we first adopt BERT and wav2vec2.0 to learn representations for the text and speech input.
% the given utterance and the label information respectively.
% % Then, it leverages the label information to locate the salient tokens/frames in the utterance via the cross-attention mechanism in each modal.
% To encourage the model to pay more attention to the emotion-related tokens in the text, it adopt a label-token attention mechanism following the encoder to capture the relations between the label and the text.
% Similarly, it also involves a label-frame attention mechanism to learn the relation between the label and the speech.
% Based on the cross-attention map of each modal, we introduce another attention mechanism to encourage the model to learn the consistance of the alignment in different modals.
% When training, besides the cross-entropy loss on each modal representation, we also introduce two auxiliary loss based on the cross-modal attention mechanism.
To locate the salient tokens/frames in the utterance, we conduct label-text/speech interactions by introducing a \textit{label-token} attention mechanism for the text and a \textit{label-frame} one for the speech, which encourages the model to pay more attention to the emotion-related tokens/frames.
Based on the above two cross-attention maps, we further introduce a label-guided attention mechanism to fuse the text and the speech.
Since the inputs of this mechanism involve emotion-related information, it is capable of aligning the text and the speech from the emotional perspective.

% In this work, we propose a novel label enhanced model for multimodal emotion recognition (LE-MER). Firstly, for the label embedding problem, we adopt a unified approach to obtain the label embeddings for both text and speech modalities, by selecting the Top-K key-words/frames as representatives for each class respectively. To this end, we necross-attentione discrete representations of speech via the quantizer module of pre-trained wav2vec2.0, then we can apply the keywords extraction algorithm upon speech frames which comes from the same class. Next, in order to learn the label-enhanced text/speech representations, we introduced a new objective based on the \textit{label-token} and the \textit{label-frame} attention matrix respectively, which encourage the emotional-relevant words/frames are weighted higher than the irrelevant ones and help the model pay attention to the emotion related words/frames. 
%  Secondly, as for the MER with label information, we further design a label-guided attentive fusion module to align the speech frames and text words by the emotion-related information, where the shared label space will serve as the bridge to integrate the emotional information from multiple cues more effectively. 

Our main contributions are summarized as follows:
1) To the best of our knowledge, this is the first work exploring a label embedding enhanced model for speech emotion recognition.
% , which maybe applied to other speech classification task.
2) We propose a novel label-guided cross-attention mechanism to fuse different modalities, which is capable of learning the alignment between speech and text from the perspective of emotional space.
3) We show the effectiveness of our method on the IEMOCAP dataset with significant improvements compared with the baseline methods.

\section{Proposed Approach}
The overview architecture of our proposed LE-MER is illustrated in Figure 2.
%In this section, we first review our proposed Multimodal Emotion Recognition with Label Embedding Attentive Model (MER-LEAM). Astransformed in Figure \ref{fig:speech_production}, 
LE-MER consists of three modules: a semantic label enhanced text encoder,  a tonal label enhanced speech encoder, and a multi-modal fusion module with label-guided cross attention.

\subsection{Semantic-label enhanced text encoder}
% \subsubsection{vanilla emotion recognition with BERT model}
%In recent years, substantial works \cite{xiong2021fusing,qiu2020pre,heusser2019bimodal} have proven that pre-trained language models, trained from large amounts of corpus, can learn universal representations that are capable of yielding good gains for downstream tasks. The widely used BERT \cite{devlin2018bert} is a bidirectional auto-encoding model that utilizes a Cloze training strategy in the pre-training phase, enabling it to efficiently extract contextual information. In this paper, we employ BERT as our text encoder. 

 Inspired by the success of Pre-trained Language Model (PLM) \cite{xiong2021fusing,qiu2020pre,heusser2019bimodal}  on numerous NLP tasks, we apply the BERT \cite{devlin2018bert} as the text encoder without loss of generality.
 For each utterance ${u_t}$, we feed it into the BERT and get the sequence representation $\mathbf{H}_{t}\in\mathbb{R}^{l_t\times d_t}$, where $l_t$ is the length of utterance and $d_t$ is the dimension.
 %Then $H_{t}$ is passed through a linear layer to get the logit for classification. The loss function can be written as 
%\begin{equation}
%\mathcal{L}^{ce}_{t}=\text{CE}(\text{Linear}(H_{t}))
%\end{equation}

 %Label embedding \cite{wang-etal-2018-joint-embedding} aims to learn representations of labels based on related information and transforms discrete categories into the semantic space. we argue that emotional words can be better extracted by fusing semantic-label embedding into text encoders. 
In order to get the emotion-aware text representation, we fuse the semantic label information into the text encoder. Firstly, we extract the keywords on the text corpus under one class in the training set to get representative textual label descriptions. Specifically, we adopt the commonly used TF-IDF \cite{beliga2014keyword} algorithm to extract the Top-K words, %which combined with the original label word will serve as the text label description for each class. 
then we feed these label descriptions into BERT, and obtain the semantic label embedding $\mathbf{L}_{t}\in\mathbb{R}^{c\times d_t}$ by averaging the token embeddings of all label descriptions, where $c$ denotes the number of classes.

After embedding both the words and the labels into a joint space, we can obtain the \textit{label-token} attention matrix $\mathbf{G}_{t}\in\mathbb{R}^{l_t\times c}$ by computing the cosine similarity between the text representation and label embedding as follows:
%For full integration of label embedding information and the sentence representation, we conduct a dot product between the sentence representation with label embedding to get the semantic similarity matrix between text and label embedding $G_{t}$ as follows

\begin{equation}
%G_{t}=\frac{H_{t}\otimes  L_{t}}{{\lVert H_{t}\rVert}_2{\lVert  L_{t}\rVert}_2} 
\mathbf{G}_{t}=\frac{\mathbf{H}_{t} \cdot \mathbf{L}_{t}^T}{{\lVert \mathbf{H}_{t}\rVert}_2{\lVert  \mathbf{L}_{t}\rVert}_2} 
\end{equation}
Then we introduced a new objective based on the \textit{label-token} interaction to encourage the emotion relevant words to be weighted higher than the irrelevant ones. Specifically, we conduct mean-pooling on the attention matrix $\mathbf{G}_t$ along the axis of sequence length, which is used as the discriminator for each class to judge the emotional relevance. Finally, we obtain the predicted logits $\mathbf{p}_g^t$ and the loss $ \mathcal{L}_g^t$:
\begin{gather}
    \mathbf{p}_g^t = \text{Softmax}(\text{Meanpooling}(\mathbf{G}_{t})) \\
    \mathcal{L}^{t}_{g}=\text{CE}(\mathbf{y},\mathbf{p}_g^t)
    %\mathcal{L}^{t}_{g}=-\sum_{i}^{N}y_{i}log(p_g)
\end{gather}
% where $\text{CE}(\cdot,\cdot)$ is the cross entropy between two probability vectors.
where $\text{CE}(\cdot,\cdot)$ refers to the cross entropy loss.

%For better convergence in the training phase, we also optimize the original logits yielded by the hidden states from the text encoder, i.e. $p_m$. Accordingly, the loss function of our semantic-label enhanced text encoder can be expressed as:
%\begin{equation}
%\mathcal{L}_t=-(\alpha \mathcal{L}^{g}_{t}+ (1-\alpha )\mathcal{L}^{ce}_{t})
%\end{equation}
%where $\alpha$ is a hyperparameter, and $y$ is the ground truth.

\subsection{Tonal-label enhanced speech encoder}
The recent success of large pre-trained models \cite{chen2022wavlm, hsu2021hubert, fu2022ufo2, mohamed2022self, baevski2022data2vec, liu2021tera, baevskivq, schneider2019wav2vec} motivates us to adopt novel, high-level features from self-supervised learning models. For the audio modality, we use wav2vec2.0\cite{baevski2020wav2vec} as our speech encoder.
%SSL pre-trained models for speech  enlighten us to utlize wav2vec2.0\cite{baevski2020wav2vec} as our speech encoder. 
%2nd stage pretraining, a strategy that has been proven\cite{chen2021exploring,gururangan2020don} to improve the performance of pre-trained model on downstream tasks, is adopted to our pre-trained speech encoder.
For each waveform of utterance, we obtain a sequence of contextualized representations from the output of wav2vec2.0 $\mathbf{H}_s\in\mathbb{R}^{l_s \times d_s}$, where $l_s$ is the number of time frames, and $d_s$ is the feature dimension.

%In vanilla speech emotion recognition, meanpooling is applied on contextualized representations to aggregate utterance level emotional information for classification, obtaining a emotion vector $e_s$. A fully-connected layer with softmax activation intakes $e_s$ and output emotion probabilities $p_s$ for emotion classification.
Similarly, we leverage the label information to obtain the emotion-aware speech representation. In order to represent the tonal label, we adopt a unified method as in text label embedding. 
That is, the key audio frames which contain representative tonal information will be extracted to generate the speech label embeddings for each class. To this end, we need to obtain the discrete representations of speech first.
The quantizer module of the pre-trained wav2vec2.0 discretizes the output of the CNN feature encoder into a finite set %of speech representation
\cite{baevski2020wav2vec}, enabling the application of key-frames extraction on the speech data.
% The quantizer module of the pre-trained wav2vec2.0 can encode local representations obtained from CNN feature extractor into discretized semantic space, enabling the application of key-frames extraction on the audio data under the same class.
In the same way, the TF-IDF is adopted to select the Top-K emotion-relevant frames under the same class. The embeddings of these emotion-relevant frames extracted from (codebook of) wav2vec2.0 will be averaged to produce the final tonal label embeddings $\mathbf{L}_{s}\in\mathbb{R}^{d_s\times c}$, where $d_s$ is the feature dimension identical to the dimension of $\mathbf{H}_s$.

Through the above way, we have embedded both the speech and the tonal label into a shared latent space, and then the \textit{label-frame} interaction matrix can be computed in the following way:

%In order to further explore and filter out emotion related information within speech contextualized representations, speech similarity matrix $G_{s}\in\mathbb{R}^{T\times c}$ is obtained by computing the cosine similarity between $H_s$ and $L_{s}$, the same process as  the calculation of text similarity matrix $G_{t}$:
\begin{equation}
\mathbf{G}_{s}=\dfrac{\mathbf{H}_s \cdot \mathbf{L}_{s}^T} {{\lVert \mathbf{H}_s\rVert}_2{\lVert \mathbf{L}_{s}\rVert}_2} 
\end{equation}
% For $ith$ row and $jth$ column element in $G_{speech}$, we can regard it as the similarity between the $jth$ time step in speech context representation $\mathcal{C}$, which is $c_j\in\mathbb{R}^{d_s}$, and $ith$ class label embedding in  $L_{speech}$, which is $L_{speech,i}\in\mathbb{R}^{d_s}$.
where $\mathbf{G}_{s}\in\mathbb{R}^{l_s\times c}$, and each element indicates the similarity between the frames and the emotion category. %In order to further filter out  emotion related information within speech contextualized representations, we introduced another objective based on \textit{label-frame}. 
In order to figure out the emotion related frames with the guidance of tonal label, we introduced another objective based on \textit{label-frame} interaction.
Specifically, the mean-pooling is conducted on the interaction matrix $\mathbf{G}_{s}$ along the frame axis to aggregate utterance level emotional correlation score. %producing the emotion similarity probability for classification with softmax activation:
\begin{equation}
    \mathbf{p}_{g}^{s}=\text{Softmax}(\text{Meanpooling}(\mathbf{G}_{s}))
\end{equation}
The \textit{label-frame} interaction based objective can be written as:
\begin{equation}
    %\mathcal{L}_{s}^{g}=-ylog(p_{s}^{g})
    \mathcal{L}_{g}^{s}=\text{CE}(\mathbf{y}, \mathbf{p}_{g}^{s})
\end{equation}
which directly encourages the speech encoder to pay more attention to the emotional frames.

\subsection{Multi-modal fusion with label-guided cross attention}
%Relying solely on either acoustic or semantic information does nis ot offer sufficient robustness in emotion recognition, which leads to increasing attention devoted to the use of multi-modal approaches. %The heterogeneity across modalities calls for research in multimodal fusion strategies, 
%The recent research mostly focus on multimodal fusion strategies, 
Multimodal fusion technology for speech emotion recognition has been widely studied in recent years, including cross-attention fusion \cite{xu2019learning}, co-attention fusion \cite{siriwardhana2020jointly}, score fusion\cite{makiuchi2021multimodal}, time synchronous and asynchronous fusion \cite{wu2021emotion}, multimodal transformer \cite{li2022context}, and etc. However, all these fusion mechanisms just devoted to aggregate word and speech embeddings, while ignored the the rich prior information contained in the emotion labels. 
We argue that the emotion labels can serve as a guidance to integrate the two modalities more efficiently.
To this end, we elaborate a novel \textit{label-guided cross-attention} to fuse multimodal emotion-related information.
%As noted in these works \cite{bahdanau2014neural,xu2015show,vaswani2017attention}, attention mechanisms can effectively capture the highly relevant semantic parts of features. However, being semantically relevant is not always consistent with emotional relevance. To this end, we elaborated a new label-guided cross-attention to fuse multimodal sentiment-related information.

The cross-attention mechanism have been proposed to capture the fine-grained interactions between the hidden representations of tokens and frames \cite{xu2019learning,zhu2022cross}:
%Firstly, we obtain the vanilla cross-attention  $A_r$ as fo
%As for the vanilla cross-attention, which formulated as the correlation map between the hidden representations of tokens and frames:
\begin{equation}
\mathbf{A}_r=\text{Softmax}(\mathbf{H}_{t}\mathbf{H}_s^{T}\mathbf{W})
\end{equation}
where $\mathbf{A}_{r}\in\mathbb{R}^{l_t \times l_s}$, and $\mathbf{W}\in\mathbb{R}^{d_s\times d_t}$ . Next, we can obtain the aligned hidden audio representation $\mathbf{H}_s^{\prime}$ by weighting $\mathbf{H}_s$ with cross-attention $\mathbf{A}_r$, and the multimodal features $\mathbf{H}_m$ can be obtained by concatenating the text representation and the aligned speech representation:
\begin{gather}
\mathbf{H}_s^{\prime}=\mathbf{A}_r\mathbf{H}_s^T,
\mathbf{H}_m=[\mathbf{H}_t, \mathbf{H}_s^{\prime}] 
 \end{gather}
%However, as mentioned above, the interaction matrix is origin from the hidden represention ...
Considering that the text and speech have been projected into the target emotional space in sections 2.1 and 2.2,
we directly multiply the \textit{label-token} 
interaction $\mathbf{G}_{t}$  and the \textit{label-frame} interaction $\mathbf{G}_{s}$  to obtain the label-guided cross-attention matrix:
\begin{equation}
\mathbf{A}_l=\mathbf{G}_{t}\cdot \mathbf{G}_{s}^{T}
\end{equation}
where $\mathbf{A}_{l}\in\mathbb{R}^{l_t\times l_s}$, and each element indicates the similarity between text tokens and speech frames from the perspective of emotional correlation. 
% That is, the shared label space serves as a bridge for the modality fusion, and the label-guided cross-attention aligns speech frames and text words toward the emotion classification goal.
% %That is, the label-guided cross-attention leverages the shared label space as a bridge to align speech frames and text words, and it has achieved emotional goal-oriented
% %To inject the emotionally relevant information into the vanilla attention matrix, we add a constraint to these two attention matrices:
% With this in mind, we propose an \textit{Attention Constraint} between the two cross-attention, to guide the vanilla attention mechanism to be aware of the emotional part by minimizing the mean squared error between them :
Compared with $\mathbf{A}_{l}$, $\mathbf{A}_{r}$ solely represents the inherent semantic relations between the text and the speech, instead of emotion specific.
To bridge the gap and integrate the emotion-aware relations into $\mathbf{A}_{r}$, we propose an \textit{Attention Constraint Module} which adopts $\mathbf{A}_{l}$ to guide $\mathbf{A}_{r}$, and we implement it with the mean squared error as follows:
\begin{equation}
%\mathcal{L}_c = \text{MSE}(A_l,A_r)
\mathcal{L}_c = ||\mathbf{A}_l-\mathbf{A}_r||^2
\end{equation}

%Finally, we feed the emotion-aware multimodal features $H_m$ generated by the label-guided cross-attention mechanisms into a linear projection to obtain the prediction result and  optimize it with the cross-entropy loss:
Finally, we aggregate the emotion-aware multimodal features $\mathbf{H}_m$ into a fixed-length vector $\mathbf{v}$ via max-pooling operation, and then feed $\mathbf{v}$ into a linear projection to obtain the prediction result and  optimize it with the cross-entropy loss:
\begin{equation}
   %\mathcal{L}^{ce}_{m}=-\sum_{t}^{N}y_{t}(\text{Softmax}(\text{Linear}(H_m)))
   \mathcal{L}_{m}=\text{CE}(\mathbf{y}, \text{Softmax}(\text{Linear}(\mathbf{v})))
\end{equation}

%In the optimization process, we optimize both the logits produced by label embedding as well as the unimodal models to ensure better convergence. The full optimization objective can be written as:
The overall loss function of the LE-MER is summarized as follows:
\begin{equation}
\mathcal{L}=\mu_{1}\mathcal{L}_{m}+\mu_{2}\mathcal{L}_{c}+ \mu_{3}\mathcal{L}_{g}^{t}+\mu_{4}\mathcal{L}_{g}^{s}
\label{eq:total_loss}
\end{equation}
where $\mu_{1}$, $\mu_{2}$, $\mu_{3}$,  and $\mu_{4}$ are hyperparameters.

% \subsection{Tables}

% An example of a table is shown in Table~\ref{tab:example}. The caption text must be above the table. Tables must be legible when printed in monochrome on DIN A4 paper; a minimum font size of 8 points is recommended.

% \begin{table}[th]
%   \caption{This is an example of a table}
%   \label{tab:example}
%   \centering
%   \begin{tabular}{ r@{}l  r }
%     \toprule
%     \multicolumn{2}{c}{\mathbf{Ratio}} & 
%                                          \multicolumn{1}{c}{\mathbf{Decibels}} \\
%     \midrule
%     $1$                       & $/10$ & $-20$~~~             \\
%     $1$                       & $/1$  & $0$~~~               \\
%     $2$                       & $/1$  & $\approx 6$~~~       \\
%     $3.16$                    & $/1$  & $10$~~~              \\
%     $10$                      & $/1$  & $20$~~~              \\
%     $100$                     & $/1$  & $40$~~~              \\
%     $1000$                    & $/1$  & $60$~~~              \\
%     \bottomrule
%   \end{tabular}
  
% \end{table}

\section{Experiments}

% Information on how and when to submit your paper is provided on the conference website.
In this section, we present the  dataset, the results compared with other state-of-the-art approaches and the related analysis.

\subsection{Dataset}
We evaluate our proposed model on the commonly used Interactive Emotional Dyadic Motion Capture (IEMOCAP) database\cite{busso2008iemocap}, which contains approximately 12 hours of audiovisual data. Among them, the text transcriptions, along with the corresponding audio, consist of five dyadic sessions where actors perform improvisations or scripted scenarios. 
% The IEMOCAP database is annotated with seven discrete categories: \textit{Neutral}, \textit{happy}, \textit{sad}, \textit{angry}, \textit{frustrated}, and \textit{excited}. 
To be consist with previous works \cite{chen2022key,9622137,hou2022multi,wu2021emotion,santoso2021speech,li2022context}, we conduct experiments on 5531 utterances from four categories: \textit{angry}, \textit{happy} (merged with \textit{excited}), \textit{sad}, and \textit{neutral}. We evaluate the model by a leave-one-session-out (5-fold) cross-validation (CV) strategy and adopt the average weighted accuracy (WA) and unweighted accuracy (UA) as evaluation metrics.

\subsection{Experimental Setup}
\textbf{Data preprocessing.}
% In terms of data augmentation on speech waveform, we follow the setting in \cite{sarma2018emotion}, conducting speed perturbation on the amplitude-modulated speech waveform with speed factors 0.9, 1.0, and 1.1. The maximum length of the input audio waveform is 20 seconds during training in order to improve training efficiency. 
For speech modality, the 80-dimensional Log Mel-spectrograms \cite{fu2021incremental} of each speech waveform are extracted by a 25ms window size with a 10ms step size and then normalized to the standardized normal distribution in utterance level. SpecAugment\cite{park2019specaugment} is also applied to the extracted acoustic feature to improve the generalization ability of the model.
% Context is the key to understanding the dialogue \cite{poria2019emotion} and can  provide contextual information as well as some additional clues to the current utterances.
% Since it is difficult to predict the content of future utterances in real-life scenarios, we only use historical utterances in this paper. 
For text modality, we use historical utterances to enhance performance, as they can provide contextual information as well as some additional clues to the current utterance \cite{poria2019emotion}. Specifically, no more than ten historical utterances are spliced for each utterance and the maximum token length is limited to 150.

\noindent\textbf{Settings.}
% For pre-trained speech e oder, we tried two wav2vec2.0-base models: wav2vec2.0-conformer\cite{zhang2020pushing} and UFO2\cite{fu2022ufo2}. Wav2vec2.0-conformer is a vtoav2vec2.0, which replaces the Transformer block with Conformer block. UFO2 is a variant of wav2vec2.0-conformer as it supports unified pretraining for both online and offline speech encoding and decoding. 2nd stage is applied on the pre-trained speech encoder, 200th checkpoint during training will be the final pre-trained speech encoder.
The pre-trained wav2vec2.0-conformer-BASE\cite{zhang2020pushing} and BERT-BASE \cite{devlin2018bert} model are employed as our speech and text encoder,  respectively. Following \cite{pepino2021emotion, chen2021exploring, wang2021fine, mohamed2022self}, wav2vec2.0 is pre-trained on 960h LibriSpeech dataset. In addition, 2nd stage pretraining is applied to the pre-trained wav2vec2.0-conformer on the training set.
% The pre-trained BERT-BASE \cite{devlin2018bert} model are employed 
% as our text encoder.
We adopt Adam as our optimizer with a warm-up of 8000 steps and set the learning rate to $10^{-6}$ for the wav2vec2.0 and $5\times10^{-6}$ for BERT model, while the batch size for training is 16. As for hyperparameters in Eq~(\ref{eq:total_loss}), we set $\mu_{1}$, $\mu_{2}$, $\mu_{3}$, and $\mu_{4}$ to 1, 0.5, 0.2, and 0.2 empirically\footnote{The source code  will be publicly available at: \url{https://github.com/Digimonseeker/LE-MER}.}.

\noindent\textbf{Baselines.}
%\subsection{Baselines}
We compare our proposed  LE-MER  with several baselines: 
\cite{chen2022key,9622137,li2022context} 
adopted cross-attention mechanism to fuse multimodal information, where a modified key-sparse attention is proposed in \cite{chen2022key}. 
Hou et al.\cite{hou2022multi} proposed a self-guided modality calibration network to achieve alignment between audio and text modalities.
Wu et al.\cite{wu2021emotion} proposed a two-branch neural network to capture correlations between multimodalitly from both word level and utterance level. 
%Santoso  et al.\cite{santoso2021speech} proposed to use the combination of a self-attention mechanism and a word-level confidence measure (CM) to mitigate the effects of speech recognition errors of Automatic Speech Recognition (ASR) systems in MER.
Santoso  et al.\cite{santoso2021speech} proposed to use the combination of a self-attention mechanism and a word-level confidence measure (CM) to mitigate the errors in MER produced by ASR system.

\subsection{Main Results}
% To evaluate the performance of our model, we conducted unimodal and multimodal experiments on the IEMOCAP dataset.
\textbf{Unimodal Results.} 
As we can observe from Table \ref{tab:Unimodal Results}, the performance of the text encoder improves significantly after integrating historical utterances (A2), proving that historical utterances can provide additional cues to support the current utterances.
% To compare with \cite{wang2018joint}, we also conducted experiments using the LEAM framework (B3). We can find that both WA and UA of the model decrease after using the LEAM framework, implying that the $G_t$ matrices are already highly information-intensive and can be used directly for classification without further fusion.%其他实验对比
In addition, we also investigate the effects of different initialization for text label embeddings (A3-A5). We can observe that A4 achieves superior performance than A3, while the best results are achieved by A5, which substantiates that our keyword initialization scheme yields a more effective and representative label embeddings than the others.
For speech modality, 2nd stage pretraining before finetuning can improve WA and UA by at least absolute 1.8 percent (B2 vs. B1).
Moreover, speech label embeddings with all three types of initialization (B3-B5) bring varying contributions compared to B2.
Some potential prior information from BERT embedding (B4) boosts the performance compared with B3. By getting rid of the shackles of the modality gap and benefiting from the discretized tonal label generated from pre-trained quantizer of wav2vec2.0, B5 makes a further improvement (B5 vs. B4) and achieves the best result.

% \begin{table}[tb]
% \vspace{-0.5em}
%   \centering
%   \caption{Comparison of our unimodal results on IEMOCAP dataset where "LE" denotes label embedding.}
  
%   \label{tab:Unimodal Results}
%   \scalebox{0.75}{
%    \begin{tabular}{c|rlcc}
%     \hline
%     Sys. & Model      &       & WA(\%) & UA(\%) \\
%     \hline
%     A1    & \multicolumn{2}{l}{BERT} & 67.34 & 67.66 \\
%     A2    &       & + historical utterances & 77.46 & 78.38 \\
%     A3    &       & + historical utterances + LE (random init) & 77.51 & 78.52 \\
%     A4    &       & + historical utterances + LE(label words init) & 78.03 & 78.88 \\
%     A5    &       & + historical utterances + LE(TF-IDF init)  & \textbf{78.11} & \textbf{78.92} \\
%     \hline
%     B1    & \multicolumn{2}{l}{wav2vec2.0} & 73.92 & 74.48 \\
%     B2    &       & + 2nd stage & 75.73 & 76.44 \\
%     B3    & \multicolumn{1}{l}{     } & + 2nd stage + LE (random init) & 76.20 & 76.80 \\
%     B4    & \multicolumn{1}{l}{      } & + 2nd stage + LE (BERT embedding init) & 76.48 & 77.14 \\
%     B5    &       & + 2nd stage + LE (codebook init) & \textbf{76.74} & \textbf{77.74} \\
%     \hline
%     \end{tabular}%
%     }

%   \vspace{-2.0em}
% \end{table}%
\begin{table}[tb]
\vspace{-0.5em}
  \centering
  \caption{Comparison of our unimodal results on IEMOCAP dataset where ``LE" denotes label embedding.}
  \vspace{-1.0em}
  \label{tab:Unimodal Results}
  \scalebox{0.75}{
   \begin{tabular}{c|lcc}
    \hline
    Sys. & Model      & WA(\%) & UA(\%) \\
    \hline
    A1    & BERT & 67.34 & 67.66 \\
    A2    & \quad + historical utterances & 77.46 & 78.38 \\
    A3    & \quad + historical utterances + LE (random init) & 77.51 & 78.52 \\
    A4    & \quad + historical utterances + LE (label words init) & 78.03 & 78.88 \\
    A5    & \quad + historical utterances + LE (TF-IDF init)  & \textbf{78.11} & \textbf{78.92} \\
    \hline
    B1    & wav2vec2.0 & 73.92 & 74.48 \\
    B2    & \quad + 2nd stage & 75.73 & 76.44 \\
    B3    & \quad + 2nd stage + LE (random init) & 76.20 & 76.80 \\
    B4    & \quad + 2nd stage + LE (BERT embedding init) & 76.48 & 77.14 \\
    B5    & \quad + 2nd stage + LE (codebook init) & \textbf{76.74} & \textbf{77.74} \\
    \hline
    \end{tabular}%
    }

  \vspace{-2.0em}
\end{table}%

\noindent\textbf{Multimodal Results.} In Table \ref{tab:multimodal result}, we compare our multimodal results on the IEMOCAP dataset with the existing state-of-the-art methods, which share the same setting of data preprocessing with us, for a fair comparison.
It shows that our proposed approach  achieves state-of-the-art results compared to the others in terms of both WA and UA.
Here we also present the result obtained by the score fusion scheme, which simply sums the logits produced by two unimodal models for predictions.
This straightforward scheme can achieve favorable result that outperforms all the baselines, indicating that our label embeddings can facilitate unimodal encoders to locate the salient tokens/frames relevant to the specific emotion, thus yielding better result.
Compared with methods based on attention mechanism, such as \cite{li2022context,chen2022key}, our model achieves superior performance, which proves that our proposed label-guided attentive fusion module can serve as a bridge to leverage cues from multimodality to integrate emotional information more effectively. 

\begin{table}[htbp]
\vspace{-0.5em}
  \centering
  \caption{
  Comparison of our multimodal results with previous works on IEMOCAP dataset.
  }
  \vspace{-1.0em}
  \scalebox{0.8}{
    \begin{tabular}{l|rcc}
    
    \hline 
         Model &        & WA(\%)    & UA(\%) \\
    \hline
    Chen et al.\cite{chen2022key}  &       & 74.30 & 75.30 \\
    Chen et al.\cite{9622137}  &       & 74.92 & 76.64 \\
    Hou et al.\cite{hou2022multi}  &       & 75.60 & 77.60 \\
    Wu et al.\cite{wu2021emotion}  &       & 77.57 & 78.41 \\
    Santoso  et al.\cite{santoso2021speech}  &       & 78.40 & 78.60 \\
    Li et al. \cite{li2022context} &       & 80.36 & 81.70 \\
    \hline
    Our Score Fusion &       & 81.32 & 82.18 \\
    % Cross-Attention &       &       &  \\
    \hline
    Ours  &       & \textbf{82.40} & \textbf{83.11} \\
    \hline
    
    \end{tabular}%
    }
  \label{tab:multimodal result}%
  \vspace{-2.0em}
\end{table}%

\subsection{Discussion}
\begin{figure}[htb]
\vspace{-1.0em}
  \centering
  \includegraphics[width=0.6\linewidth]{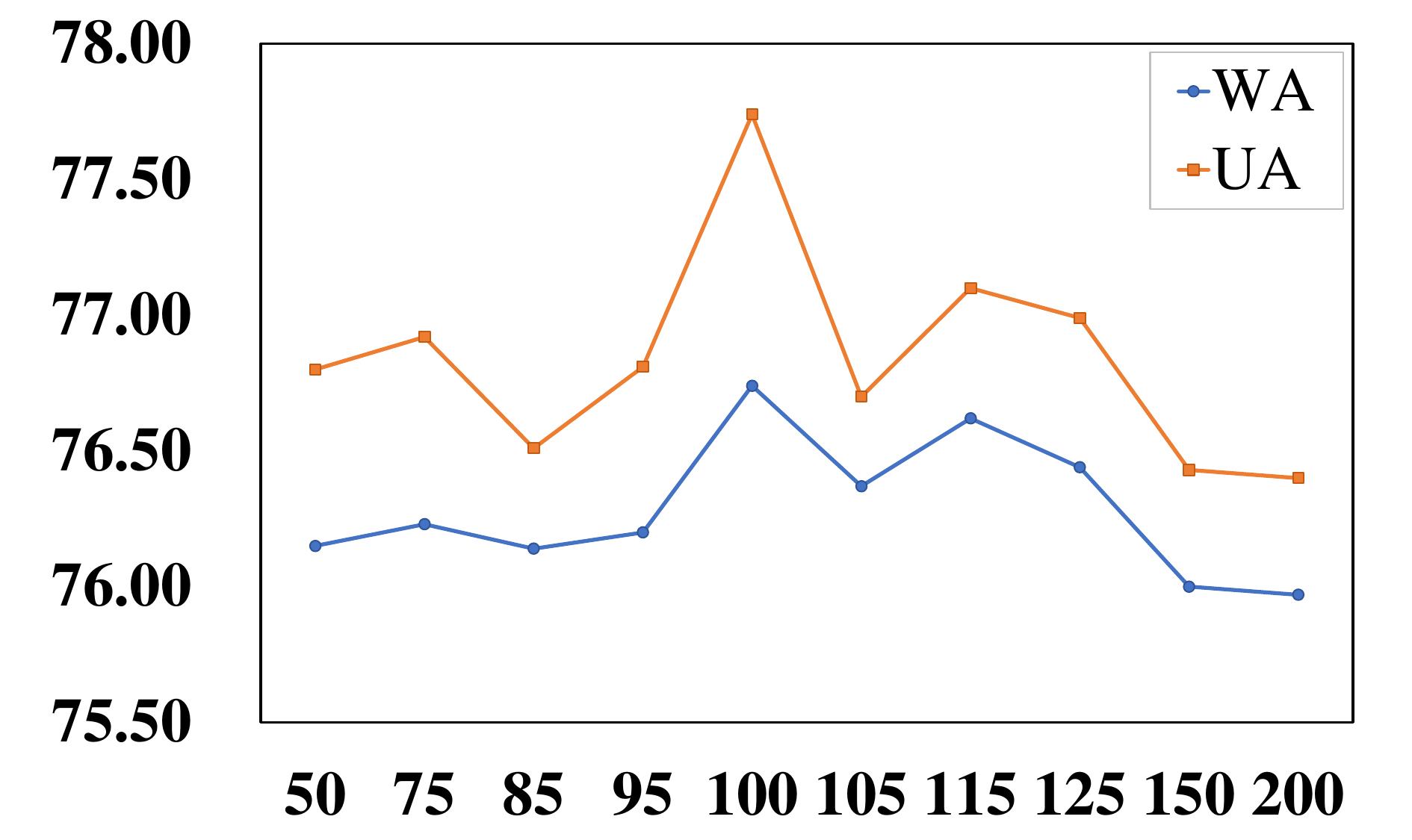}
  \vspace{-1.0em}
  \caption{Effect of K for speech label embeddings initialization.}
  \label{fig:TOPK-speechlabelembedding}
  \vspace{-1.0em}
\end{figure}

\textbf{Hyper-parameter Tuning of $K$.} 
% For the sake of exploring the optimal $K$ for wav2vec2.0 codebook initialization, we grid search $K$ from 50 to 200, the results are shown in Figure \ref{fig:TOPK-speechlabelembedding}.
To explore the optimal number of frames for the tonal label, we conduct a grid search to obtain its value.
As shown in Figure \ref{fig:TOPK-speechlabelembedding}, when $K$ is larger than 100, both WA and UA decrease to some extent, implying that label embeddings with larger $K$ contain some redundant information that is irrelevant with the corresponding type of emotion. Vice versa, label embeddings with smaller $K$ lack enough emotion-related information, causing performance degradation.
Therefore, we set $K$ to 100 for the tonal label.
% For text modality, the best $K$ is set to 9.
% We omit the process of searching for $K$ here for simplicity.
As for the semantic label, we explore the optimal $K$ with the same method, and the best $K$ is set to 9.
For sake of repetition, we omit the process here.

\noindent\textbf{Ablation study of Attention Constraint.} In this section, we further explore how to utilize the label-guided attention matrix to improve the model performance in terms of modality fusion, and we present the corresponding results in Table \ref{tab:Ablation study}. 
% We first add label-guided attention matrix to the vanilla attention matrix as a new attention matrix to fuse the multimodal information, the result is presented in Table \ref{tab:Ablation study}. We can find that a subtle decrease in the model's performance occurs.
% In addition, we performed the multimodal fusion with only label-guided attention matrix or the vanilla attention matrix. We can easily note that the performance of the model decreases with either the vanilla attention matrix or the label-guided attention matrix removed, which indicates that both matrices provide vital information assistance to the MER task.
A subtle decrease can be observed from C1 to C2 in terms of UA, revealing the superiority of the attention constraint scheme against simple summation of label-guided attention $\mathbf{A}_l$ and vanilla attention $\mathbf{A}_r$. We can attribute this performance gap to adopting attention constraint, as it provides a more delicate way to craft multimodal features
$\mathbf{H}_m$ to be label-aware under the supervision of $\mathbf{A}_l$. 
Furthermore, we perform multimodal fusion with only $\mathbf{A}_l$ (C3) or $\mathbf{A}_r$ (C4). Further performance degradation can be observed by comparing either C3 or C4 with C2, validating the necessity of the interaction between $\mathbf{A}_l$ and $\mathbf{A}_r$.


\begin{table}[htbp]
  \centering
  \vspace{-0.5em}
  \caption{Results of comparison between different fusion methods utilizing label-guided attention $\mathbf{A}_l$ and vanilla attention $\mathbf{A}_r$}
    \scalebox{0.8}{
    \vspace{-1.0em}
    \begin{tabular}{c|lcc}
    \hline
    Sys.  & Model     & WA(\%)  & UA(\%) \\
    \hline
    C1    & Attention Constraint  & \textbf{82.40} & \textbf{83.11} \\
    C2    & $\mathbf{A}_r + \mathbf{A}_l$ & 82.39  & 82.75  \\
    C3    & only $\mathbf{A}_l$ & 81.29  & 81.37  \\
    C4    & only $\mathbf{A}_r$ & 81.08  & 81.68  \\
    \hline
    \end{tabular}%
    }
  \label{tab:Ablation study}%
  \vspace{-1.0em}
\end{table}%

\begin{figure}[htb]
\vspace{-1.0em}
  \centering
  \includegraphics[width=0.9\linewidth]{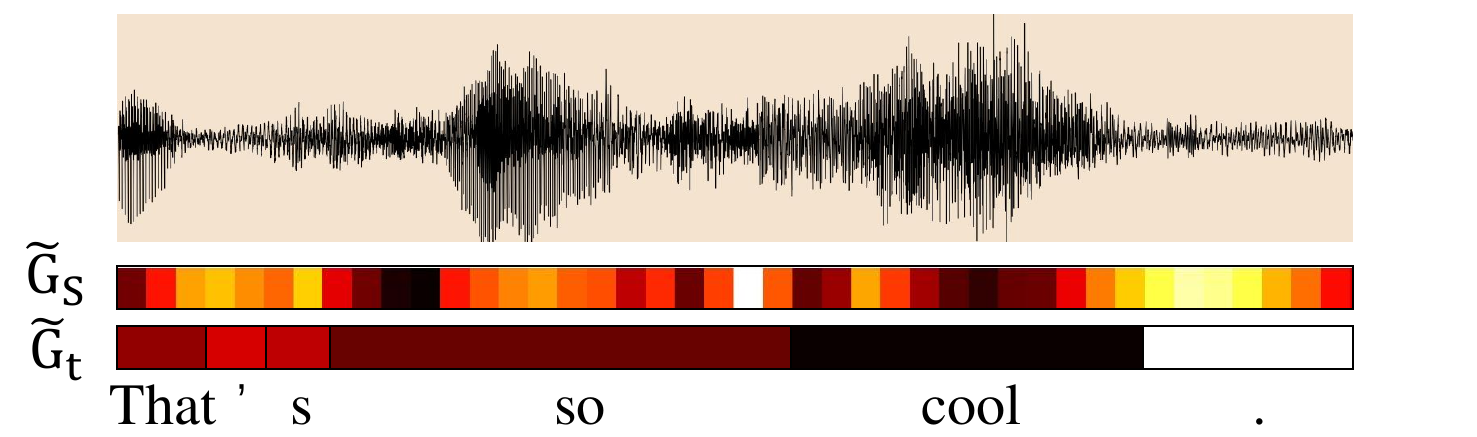}
  \vspace{-0.5em}
  \caption{Visualization of $\widetilde{\mathbf{G}}_s$ and $\widetilde{\mathbf{G}}_t$.}
  \label{fig:attention_visualization}
  \vspace{-2.0em}
\end{figure}

\noindent\textbf{Attention Visualization.} To demonstrate the effectiveness of our unimodal label embeddings, we perform the visualizations of both $\mathbf{G}_s$ and $\mathbf{G}_t$ on one utterance in IEMOCAP and present them in Figure \ref{fig:attention_visualization}.
% In this example, waveform is aligned with the tokenized text, and both $\mathbf{G}_s$ and $\mathbf{G}_t$ have been averaged over the class dimension to generate one dimension attentions $\widetilde{\mathbf{G}}_s$ and $\widetilde{\mathbf{G}}_t$.
In this example, waveform is aligned with the tokenized text, and both $\mathbf{G}_s$ and $\mathbf{G}_t$ have been averaged over the class dimension to generate the vectors $\widetilde{\mathbf{G}}_s$ and $\widetilde{\mathbf{G}}_t$.
% It is worth mentioning that the $\widetilde{\mathbf{G}}_s$ assigns higher attention weights to emotion-related speech segments, while $\widetilde{\mathbf{G}}_t$ has higher weights on some emotional words, such as “cool”.
Results show that the $\widetilde{\mathbf{G}}_s$ assigns larger attention weights to emotion-related speech segments, while $\widetilde{\mathbf{G}}_t$ has higher weights on some emotional words, such as “cool”.
This reveals that our unimodal label embeddings can effectively guide the encoders to focus on emotion relevant information from the input.

\section{Conclusions}
In this paper, we presented LE-MER, a novel multimodal fusion framework for speech emotion recognition, which takes advantage of the both the textual and speech label information to extract the emotional token and frames, respectively. By mapping the  speech and text representations to a common emotional space, we can learn the alignment between the text words and speech frames and fuse the emotional information more efficiently. Experimental results on the public IEMOCAP dataset demonstrated the superior performance of LE-MER and the importance of each component. 
% In the future, we will explore how to integrate label embedding with other modalities and more effective modality fusion framework based on label information.
In the future, we will explore how to extend this method to other speech tasks.

\bibliographystyle{IEEEtran}
\bibliography{mybib}

\end{document}